\documentclass[11pt]{IEEEtran}
\IEEEoverridecommandlockouts
\usepackage{url}
\usepackage[ampersand]{easylist}
\usepackage{graphicx, lipsum}
\usepackage{subcaption}
\usepackage{siunitx}
\usepackage{booktabs}
\usepackage[font=small,labelfont=bf]{caption}
\DeclareCaptionLabelFormat{cont}{#1~#2\alph{ContinuedFloat}}
\usepackage[figuresright]{rotating}
\usepackage{makebox}
\usepackage[table]{xcolor}
\usepackage{calc}
\usepackage{longtable}
\usepackage{changepage}
\usepackage[nottoc,notlof,notlot]{tocbibind} 
\usepackage{enumerate}
\usepackage{amsmath}
\usepackage{amssymb}
\usepackage{supertabular} 
\usepackage{enumitem}
\usepackage{multicol}
\usepackage{colortbl} 
\usepackage{nicefrac}
\usepackage{booktabs}
\usepackage{subcaption}
\usepackage{multirow}
\usepackage{mathtools}
\usepackage{enumitem}

\newcommand\addtag{\refstepcounter{equation}\tag{\theequation}}
\newcommand{\etal}{\textit{et al}.}

\newcolumntype{M}[1]{>{\centering\arraybackslash}m{#1}}
\newcolumntype{N}{@{}m{0pt}@{}}
\usepackage{cite}
\usepackage{amsmath,amssymb,amsfonts}
\usepackage{algorithmic}
\usepackage{graphicx}
\usepackage{textcomp}
\usepackage{xcolor}
\def\BibTeX{{\rm B\kern-.05em{\sc i\kern-.025em b}\kern-.08em
		T\kern-.1667em\lower.7ex\hbox{E}\kern-.125emX}}

\DeclarePairedDelimiter\norm{\lVert}{\rVert}%
\DeclarePairedDelimiter\abs{\lvert}{\rvert}%
\DeclareGraphicsExtensions{.pdf,.png,.jpg}

\makeatletter
\let\oldabs\abs
\def\abs{\@ifstar{\oldabs}{\oldabs*}}
\let\oldnorm\norm
\def\norm{\@ifstar{\oldnorm}{\oldnorm*}}
\makeatother

\begin{document}

\title{BRIEF Parameter Analysis and the Colour Extension for Texture Classification}

\author{Qinhao Wu,
        Nora Al-Garaawi,
        Tim Morris,
        
        % <-this % stops a space
%\thanks{This research was supported by the National Natural Science Foundation of China, Young Scientists Fund, 72201193, “Study on the multi-attribute assessment of safety risk of Arctic navigation using dynamic belief rule-base inference methodology.” (Corresponding author: Shuaiyu Yao)}% <-this % stops a space
\thanks{Qinhao Wu is with the Leiden Academic Centre Drug Research, Leiden Unviersity, Leiden, 2333 AL, Netherlands (e-mail: q.wu@lacdr.leidenuniv.nl).}
%\thanks{Shuaiyu Yao is with the Department of Control Science and Engineering, College of Electronic and Information Engineering, Tongji University, Shanghai, 201804, China, and also with the Shanghai Institute of Intelligent Science and Technology, Tongji University, Shanghai, 200092, China (e-mail: alvinyao@tongji.edu.cn). }% <-this % stops a space
\thanks{Nora Al-Garaawi is with the Department of Computer Science, Faculty of Education for Girls, University of Kufa, Kufa, 540011, Iraq.}% <-this % stops a space
%\thanks{Jian-Bo Yang is with the Alliance Manchester Business School, The University of Manchester, Manchester, M13 9PL, United Kingdom.}% <-this % stops a space
%\thanks{Shengzhao Chen is with the Faculty of Science, University of Nottingham, Nottingham, NG7 2RD, United Kingdom.}% <-this % stops a space
%\thanks{Taimin Lin is with the School of Political Science \& International Relations, Tongji University, Shanghai, 201804, China.}% <-this % stops a space
\thanks{Tim Morris is with the Department of Computer Science, The University of Manchester, Manchester, M13 9PL, United Kingdom.}% <-this % stops a space

}
	
\maketitle

\begin{abstract}
	BRIEF (Binary Robust Independent Elementary Features) is widely used in image matching and related fields. In this paper, we analyse and adapt BRIEF in texture classification systematically. BRIEF can be seen as a generalized version of Local Binary Pattern (LBP) which has achieved remarkable performance in texture classification. Based on this observation, we analyse the behaviour or BRIEF in the texture classification problem. We found that, given a fixed patch size, BRIEF achieves a better classification result when the intensity tests are overloaded.
	
	Also, a colour extension of BRIEF (C-BRIEF) is proposed. It can wrap up the texture and colour information at the same time but with a shorter histogram length, compared to the other descriptors. It shows a competitive result with FV-CNN, tested using RoTeB. It achieved the best classification score in the CUReT, KTH-TIPS2-b, and Outex\_TC12 datasets. At the same time, it is a quick and light-weight descriptor.

\end{abstract}


\section{Introduction}
Texture is a key source of information in the computer vision area. It is used in several tasks, such as object detection, image matching and classification. Texture assessment has until recently been defined by the measurement of local intensity changes and there is a large number of hand-crafted texture descriptors. Nowadays, with the development of deep texture representations~\cite{cimpoi2015deep,lin2015bilinear,andrearczyk2016using,zhang2016deep},  texture classification has been pushed to another level, showing higher accuracy and stronger robustness. This has also demonstrated the trade-off between computational complexity and recognition accuracy. Therefore, it raises a challenging problem which is can  a hand-crafted descriptor  achieve a comparable result to a deep descriptor in texture classification?

%In this paper, we mainly focus on the hand-crafted patch-based descriptor, which could be directly applied to raw pixels.
Ojala \etal~\cite{ojala1996comparative} reviewed the Local Binary Pattern (LBP) and the derived family~\cite{liu2017local}, we found that the intensity test number is the most crucial parameter in texture classification. From the original LBP, uniform LBP~\cite{ojala2002multiresolution}, and the later methods, Binary Rotation Invariant and Noise Tolerant (BRINT)~\cite{liu2014brint} and Median Robust Extended LBP (MRELBP)~\cite{liu2016median}, the intensity test number increased with the geometric pattern change. This phenomenon gives us the motivation to find a texture descriptor to contain more intensity tests without the pattern design.

As one of the potential solutions, Binary Robust Intensity Elementary Feature (BRIEF)~\cite{calonder2010brief} was suggested. It  distributes intensity tests by using the Gaussian distribution. It has been widely used as a texture descriptor in image matching and related fields. However, it has also shown the capability to be used as a patch-based descriptor in texture classification applications~\cite{mohammed2014optic} and face classification applications~\cite{al2021brief}. Moreover, the Gaussian distribution pattern can be seen as a generalized version of the LBP pattern, which can easily fit in more intensity tests in a given sampling patch. 

In this paper, we will show the parameter behaviour of BRIEF in the texture classification problem. Also, the colour extended BRIEF (C-BRIEF) will be introduced. Evaluation will show it achieves a competitive result compared with the other methods, including  robustness to illumination and scale change, and resistance to noise, whilst retaining low computational complexity. Followed by that, the potential maximum need for the intensity tests will be carried out.

\section{Related Work}\label{sec:rw}
Texture is generally assessed by measuring local intensity variations. There are two general approaches: 1. extract and encode the intensity differences in a local patch, such as the LBP family of methods~\cite{ojala2002multiresolution, ahonen2009rotation, liu2014brint, liu2016median} etc.; and 2. gather textons using filter banks, such as MR8~\cite{varma2005statistical}, and deep filter banks~\cite{cimpoi2015deep,lin2015bilinear,andrearczyk2016using,zhang2016deep,fujieda2017wavelet} etc. 

As one of the popular texture measurements, LBP and its family achieved  great success in texture classification and have been widely used. LBP$_{r,p}$ was defined as a circular shape pattern~\cite{ojala2002multiresolution}. Moreover, with the introduction of ``uniform'' patterns and the ROR operator, LBP$^{riu2}_{r,p}$ was presented and achieved rotation invariance. Specifically, the ROR operator only accumulated the local patches matching the ``uniform'' ones to form a shorter histogram of the salient information. The ROR operator was extended in most of the succeeding LBP methods. Based on the completed ``uniform'' pattern, LBP Histogram Fourier features (LBP-HF)~\cite{ahonen2009rotation} was developed. LBP-HF transformed the local texture from the rotation pattern to the occupied neighbour's phase, so that it achieved a better resistance against rotation than LBP$^{riu2}_{r,p}$.  

Then, with multiple patterns proposed, the texture histogram has been widely increased. BRINT~\cite{liu2014brint} and MRELBP~\cite{liu2016median} were two of the most recent methods in the LBP family. BRINT was extended from Completed LBP (CLBP)~\cite{guo2010completed}, which offers LBP-Sign, LBP-Magnitude, and LBP-Centre. The CLBP histogram was generated by the joint histograms of both components. Combining the concept of CLBP$^{riu2}_{r,p}$ and merging $8q\ (q\in\mathbb{N}^{+})$ pixels into the average pattern, BRINT offered better performance than CLBP. We compare BRINT\_CS\_CM against our method. On the other hand, MRELBP  cascaded multi-radii of $3\times3$ patches to concatenate the texture histogram. Despite these LBP family methods  using ROR to shrink the histogram's dimension, the descriptor grew longer with the need to represent multi-scale patterns.

BRIEF~\cite{calonder2010brief} was first introduced as a feature point descriptor for image matching, succeeding from Scale-Invariant Feature Transform (SIFT)~\cite{lowe2004distinctive} and Speeded Up Robust Features (SURF)~\cite{bay2006surf}. We observe that the BRIEF pattern could be seen as the general version of LBP$_{r,p}$. However, as it was originally designed for image matching, it used the Gaussian random pattern for selecting points and the Hamming distance to accelerate the comparisons and boost the pattern matching accuracy~\cite{calonder2010brief,calonder2012brief}. Afterwards, it was tested and used as the feature extraction for texture classification~\cite{mohammed2014optic} and segmentation~\cite{mohammad2017binary}. However, these works only tested limited combinations of the BRIEF parameters. We will show the parameter analysis of BRIEF in  texture classification to understand the parameters' behaviour and achieve a better representation of different textures.

\section{BRIEF as Texture Measure}\label{sec:methods}
\subsection{BRIEF Overview}

The BRIEF descriptor~\cite{calonder2010brief, calonder2012brief} is defined as a binary string formed by $n_d$ number of intensity tests, resulting in a $2^{n_d}$ length histogram. The locations  of pixels contributing to the tests are defined as $(X,Y)\sim i.i.d.\hspace{0.3em} Gaussian(0,\nicefrac{S^2}{25})$, where $X$ and $Y$ are the two location sets. As the original geometry is designed for image matching, it needs to be tested under the settings of texture classification in Section~\ref{sec:geometry}. Therefore, given the selected pixels $x$ and $y$ to the intensity test $\tau$, defined in Equation~\ref{brieft}, we would get a $n_d$ length binary string by the order of $(X, Y)$. The decimal BRIEF value is encoded by Equation~\ref{briefr}. 
\[
\tau(\textbf{p};x,y):=\left\{
\begin{array}{lcl}
1&,&if\hspace{0.5em}\textbf{p}(x)<\textbf{p}(y),\\
0&,&otherwise.\\
\end{array}
\right.\addtag \label{brieft}
\]
\[
f_n(\textbf{p}):=\sum\limits_{1\leq i\leq n}2^{i-1}\tau(\textbf{p};x_i,y_i).\addtag \label{briefr}
\]

Notice that, in the original paper~\cite{calonder2010brief}, the input image is  Gaussian blurred to reduce the random noise before the intensity comparisons are performed. However, as the blurring process will compromise the texture appearance, we  calculate the intensity tests using raw pixels. In order to eliminate the effect of noise, we  calculate the noise level $\theta$ as a threshold prior to the intensity test to control the noise effect~\cite{mohammed2014optic}. Equation \ref{brieft} is modified into
\[
\tau(\textbf{p};x,y):=\left\{
\begin{array}{lcl}
1&,&if\hspace{0.5em}\textbf{p}(x)-\textbf{p}(y)<\theta,\\
0&,&otherwise.\\
\end{array}
\right. \addtag \label{eq:brieftest}
\] 
\[
\begin{array}{rcl}
\theta& =& 3\cdot\sigma_{noise}\\ 
&=&3\sqrt{(G(I)-I)^2 } \addtag \label{eq:noise}
\end{array}
\]
where $G(\cdot)$ represents the Gaussian blur with the variance value of 2~\cite{calonder2012brief}, and $I$ denotes the input image. Through our result, we found that the new threshold $\theta^{\prime}$ gave a better result, which is shown in Equation~\ref{eq:noiseprime}.
\[
\begin{array}{rcl}
\theta^{\prime}& =& 3\cdot\sigma_{noise}+\mu_{noise}\\ 
&=&3\sqrt{(G(I)-I)^2 }+\nicefrac{\sum_{\textbf{p}}\norm{G(I)-I}}{n_p} 
\end{array}
\addtag \label{eq:noiseprime}
\]
where $n_p$ is the total  number of pixels. The comparison result will be shown in Section~\ref{sec:noise}.

\subsection{Colour-Extended BRIEF}
The straightforward approach to applying a hand-crafted descriptor to colour images is to extract features from each channel, shown in Section~\ref{sec:rw}. However, this method would cost triple the computational time and memory. Therefore, a multi-channel texture descriptor is needed, so that we can build up the connection between channels and give the abstract textural representation, encoded with the colour information. It will give a richer texture description compared with the greyscale method. 
\begin{figure}[t!]
	\centering\includegraphics[width = 0.5\columnwidth]{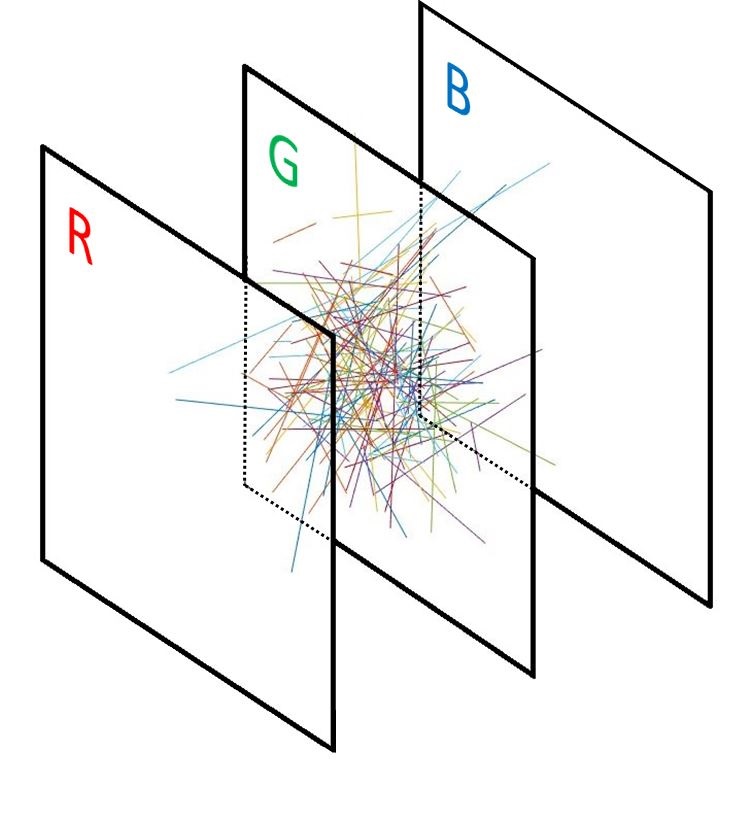} 
	\caption{C-BRIEF descriptor example with 128 tests.}\label{ss} 
\end{figure} 

To extend this method to represent colour texture, we added the colour channel location set $Z$ to the original geometry. The descriptor example is shown in Figure~\ref{ss}. However, there are two options: 
\[Z_1 \sim i.i.d.\ Uniform(3),\]
 and 
\[Z_2 \sim i.i.d.\ Gaussian(0,1).\] 

As the uniform random option would select equally from all channels, with the RGB image input, the Gaussian kernel would make it more focused on the green channel, since it contains more shape information~\cite{anderson1996proposal, mohammed2014optic}. Even though we could directly use the same distribution of $X$ and $Y$, it would create a similar probability near the centre, resulting in a mild focus on the green channel.

Furthermore, C-BRIEF can be degenerated into the normal BRIEF when dealing with the greyscale image, since all three channels would share the same value at the given pixel. Therefore, in the evaluation of texture classification, BRIEF would not be tested individually. The performance of BRIEF could be referred to as the result of C-BRIEF on greyscale datasets. Also, both colour texture descriptors, C-BRIEF$_{Uniform}$ and C-BRIEF$_{Gaussian}$, will be given in Section~\ref{sec:evaluation}.

\subsection{Texture Classification}
\subsubsection{Parameter Settings}\label{sec:optsetting}
The input image is separated into patches with size $S\times S$ with a one-pixel stride and the texture of each patch is assessed by Equation \ref{eq:brieftest} resulting in the BRIEF map. Assembled by Equation \ref{briefr}, the histogram of the given image $I$ is generated. In order to represent the  texture  of each class, the model histogram $H_m$ is calculated by the average of the given training histograms.

For the parameter analysis of BRIEF, we choose to use pixel-wise classification. The $\chi^2$ distance~\cite{varma2005statistical} between the local sub-histogram $H$ and the model is calculated as 
\[
\chi^2_{H,H_m} = \frac{1}{2}\sum_{i=0}^{L-1}\frac{\left[H(i)-H_{m}(i)\right]^2}{H(i)+H_{m}(i)} \addtag \label{chisquare}
\]
where $L$ is the full length of the generated BRIEF histogram, which is $2^{n_d}$. Furthermore, the classifying pixel's local sub-histogram $H$ is formed by its central and neighbour patches. After comparing the sub-histogram with all model histograms, the prediction label of the given pixel is suggested by the closest model histogram. 

\subsubsection{Evaluation Settings}\label{sec:evasetting}
According to RoTeB~\cite{liu2016evaluation}, the Nearest Neighbour Classifier (NNC) based on $\chi^2$ distance measurement is used for evaluation. The training set $\textbf{H}$ is given by $n$ histograms from each training image. By comparing the test histogram with $\textbf{H}$ individually, the predicted label would be given by the closest histogram $H_i\in \textbf{H}$. 

%%table for the geometry result
\begin{table*}[t!]
	\centering
	\small
	\caption{Classification Accuracy of Different Standard Deviation Denominator}
	\label{tab:more}
	%\begin{adjustwidth}{-0.5cm}{}
		\begin{tabular}{l|cccccc}
			\hline\toprule
			&\multicolumn{6}{c}{Standard Deviation Denominator}\\ 
			
			&&&&&&\\[-10pt]
			& $3$&$5$&$7$&$10+5$&$30+5$&$60+5$\\ \midrule
			Accuracy& $ 0.63\pm 0.03$&$\mathbf{0.67\pm 0.01}$&$0.64\pm 0.01$&$0.64\pm0.02$&$0.60\pm 0.05$&$0.59\pm 0.04$\\
			\bottomrule\hline
		\end{tabular}
	%\end{adjustwidth}
\end{table*}

\section{Parameter Analysis of BRIEF}\label{sec:optimize}
\begin{figure}[h!]

		\centering
		\includegraphics[width=\linewidth]{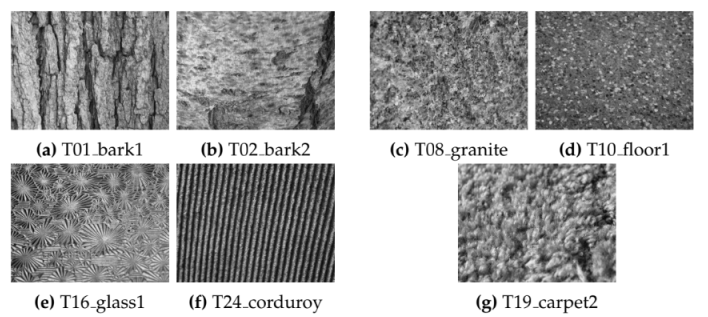}
		\caption{Example pictures of the selected textures from UIUCTex.}
		\label{fig:example_texture}
\end{figure}

In this section, we will show all individual parameter selections of BRIEF for texture classification. Our experiments are based on UIUCTex~\cite{lazebnik2005sparse}. We selected 7 difficult textural samples: T01\_bark1, T02\_bark2, T08\_granite, T10\_floor1, T16\_glass1, T24\_corduroy and T19\_carpet2, shown in Fig~\ref{fig:example_texture}.
\iffalse
\begin{table}[h!]
	\caption{Selected Class from UIUCTex}
	\label{tab:class}
	\begin{tabular}{l|cl|ll}
		\toprule
		%\multirow{2}{*}{UIUCTex}&\multicolumn{4}{c}{Directed Indication}\\
		&\multicolumn{2}{c|}{Direction}&\multicolumn{2}{c}{Directionless}\\
		\midrule
		\multirow{2}{*}{Class}&T01\_bark1& T02\_bark2&T08\_granite& T10\_floor1\\
		&T16\_glass1& T24\_corduroy&\multicolumn{2}{c}{T19\_carpet2}\\
		\bottomrule
	\end{tabular}
\end{table}
\fi
\begin{figure}[h!]
	\hspace*{\fill}
	\begin{subfigure}{0.15\textwidth}
		\centering
		\includegraphics[width=0.95\linewidth]{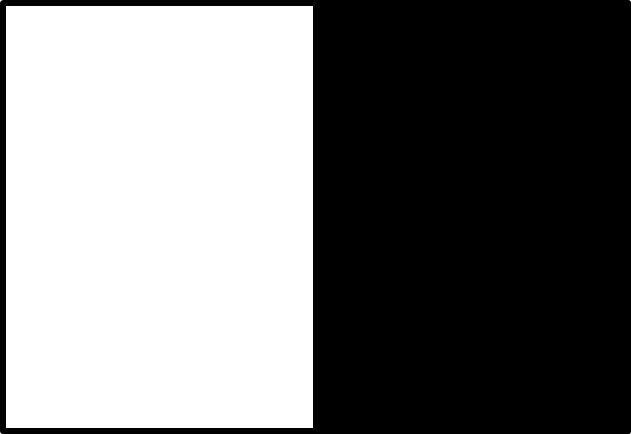}
		\caption{Mask 1}
		\label{fig:normalmask1}
	\end{subfigure}%
	\hspace*{\fill}
	\begin{subfigure}{0.15\textwidth}
		\centering
		\includegraphics[width=0.95\linewidth]{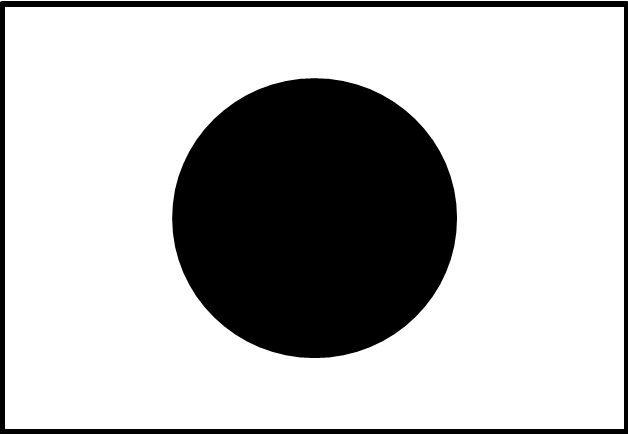}
		\caption{Mask 2}
		\label{fig:normalmask2}
	\end{subfigure}%
	\hspace*{\fill}
	\caption{Two types of test masks.}
	\label{fig:maskexample}
\end{figure}

We combined two different textures with the given mask, shown in Fig~\ref{fig:maskexample}, to create the test suites. All textural combinations are listed in Appendix Table~\ref{tab:combination}. Measured by the $\chi^2$ distance, the behaviour of each of BRIEF's parameters is shown, including the geometry pattern, noise level, patch size, and intensity test number. Therefore, based on the statistical analysis, the optimal parameter selection is identified.

\subsection{Geometry Pattern}\label{sec:geometry}

%\subsubsection{Further Analysis of \textit{G1} and \textit{O}}
The original BRIEF paper~\cite{calonder2010brief, calonder2012brief} suggested that sample point locations were generated according to a Gaussian random sampling. However, as this was proposed for image matching, it might not be suitable for texture classification. Therefore, we need to re-examine the patterns under the classification task. The original five candidate sample point distributions are:
\begin{enumerate}[itemsep=1pt]
	\addtolength{\itemindent}{1.5cm}
	\item[\textit{G1}:]  $(X,Y)\sim i.i.d.\ Gaussian(0,(\nicefrac{S}{5})^2)$; 
	\item[\textit{G2}:] $X\sim i.i.d.\ Gaussian(0,(\nicefrac{S}{5})^2),\  \\\hspace*{1.5cm} Y\sim i.i.d.Gaussian(x_i, (\nicefrac{S}{10})^2)$;
	\item[\textit{O}: ] $X=\{\forall x_i\equiv (0,0)^{\top},x_i\in X\}\\\hspace*{1.5cm} Y$ takes all possible values on a coarse polar grid containing $n_d$ points;
	\item[\textit{R1} and \textit{R2}:] $(X,Y)\sim i.i.d.\ Uniform(-\nicefrac{S}{2},\nicefrac{S}{2})$.
\end{enumerate}
\begin{figure}[h]
		\centering
		\includegraphics[width=0.95\linewidth]{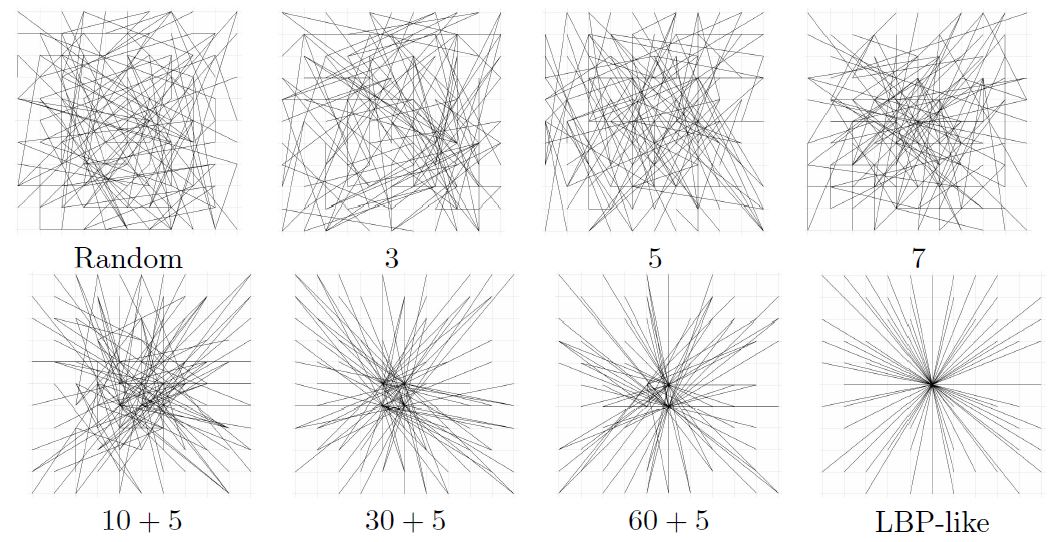}
		\caption{The example pattern from random to the concentrated generator. The number represents the denominator value. All examples' patch size is $11\times11$, filling with 128 intensity test numbers. The number under each pattern represents the standard deviation's denominator of Gaussian distribution. Specifically, the second row separates pixel set $X$ and $Y$, represented as ``denominator of $X$ + denominator of $Y$''.}
		\label{fig:geo2}
\end{figure}

Notice that \textit{O} can be seen as a similar geometry of LBP~\cite{ojala2002multiresolution}. The Gaussian random selected $Y$ can be seen as the expansion of the LBP ring with a fixed radius. Also, the centre of \textit{O} can be seen as the $X\sim i.i.d.\ Gaussian(0,\nicefrac{S^2}{\infty})$. Therefore, we can make the connection between LBP and BRIEF by replacing the standard deviation's denominator of Gaussian distribution to generate the pixel pair set $(X, Y)$. Starting from a uniform random pattern to an LBP-like pattern, we replaced the denominator with 3, 5, 7, shown in the first row of Fig~\ref{fig:geo2}. However, towards the LBP-like pattern, pixels of $X$ are gradually converged to the central pixel with the denominator from 10, 30, 60 to $\infty$, while $Y$ keeps the denominator of 5, shown in the second row of Fig~\ref{fig:geo2}. As we selected $7\times7$ patch and 9 intensity tests for the experiment, the denominator with value 60 of set $X$ would generate a relatively small standard deviation ($\sigma\approx0.1$), resulting in the central pixel being mostly selected. Therefore, it could be seen to be similar to the \textit{O} pattern. 
%% what do you mean by loosed????

The result of using the different geometric patterns is listed in Table \ref{tab:more}. We found that the LBP-like geometry pattern gave a better result with the expansion of set $X$, from column ``$60+5$'' to ``$10+5$''. However, the results of denominator 5, which is the original \textit{G1}, achieved better scores compared with the other patterns. Therefore, we continue to use \textit{G1} to generate the descriptor.

\subsection{Noise Level}\label{sec:noise}
According to Mohammed \etal~\cite{mohammed2014optic}, the original noise reduction strategy of applying the standard Gaussian filter to the image is replaced by the noise threshold $\theta_{noise}$ in Equation~\ref{eq:noise}. However, it was found that a stricter noise threshold would benefit the texture classification accuracy, resulting in a proposed threshold $\theta^{\prime}$ in Equation~\ref{eq:noiseprime}.

The stricter threshold $\theta^{\prime}$ allows fewer pixel pairs $(x,y)$ to pass the test $\tau$, resulting in a smoother histogram than that generated when using the threshold $\theta$. However, it can lead to a characteristic texture representation. As shown in Table~\ref{tab:noise}, we tested both noise thresholds using UIUCTex with $7\times 7$ patch and 9 intensity tests. It showed that $\theta^\prime$ could offer a better result. Therefore, $\theta^{\prime}$ is used in the texture classification task. 

\begin{table}[ht!]
	\centering
	\small
	\caption{Classification Accuracy Using Different Noise Thresholds}
	\label{tab:noise}
	\begin{tabular}{c|cc}
		\hline\toprule
		\multirow{2}{*}{} 	&\multicolumn{2}{c}{Noise Threshold} \\
		&$\theta$&$\theta^{\prime}$\\\midrule
		Accuracy&$0.59\pm0.01$&$0.62\pm0.03$\\
		\bottomrule\hline
	\end{tabular}
\end{table}

\subsection{Patch Size and Number of Intensity Tests}
In this section we show the BRIEF performance when changing the patch size and number of intensity tests, resulting in an optimal selection for these two parameters. In the experiment, the value of patch size $S$ ranges from 5 to 33, with step size 2; and the number of tests  $n$ is  from 1 to 20, with step size 1. The experiment is repeated ten times for each combination of parameters. The heat map of the classification result is shown in Figure \ref{fig:am}a.
\begin{figure}[ht]
	\centering 
	\ContinuedFloat*
	\includegraphics[width=0.99\columnwidth]{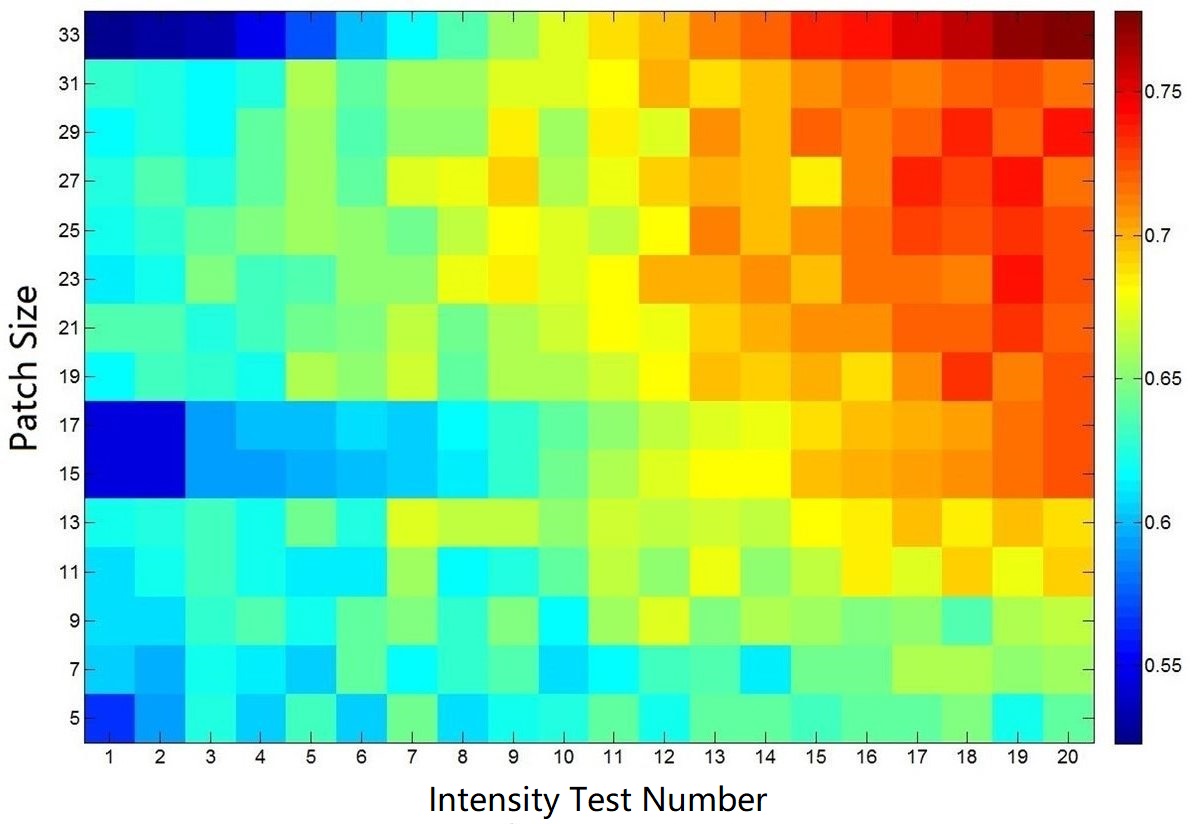}
	\caption{Accuracy heat map of the combination of the patch size, $S\in[5,33]$ with stride 2, and intensity test number, $n_d\in [1,20]$ with stride 1.}\label{fig:am}  
\end{figure}
\begin{figure}[ht]
	\centering
	\ContinuedFloat
	\includegraphics[width=\columnwidth, height = 0.55\columnwidth]{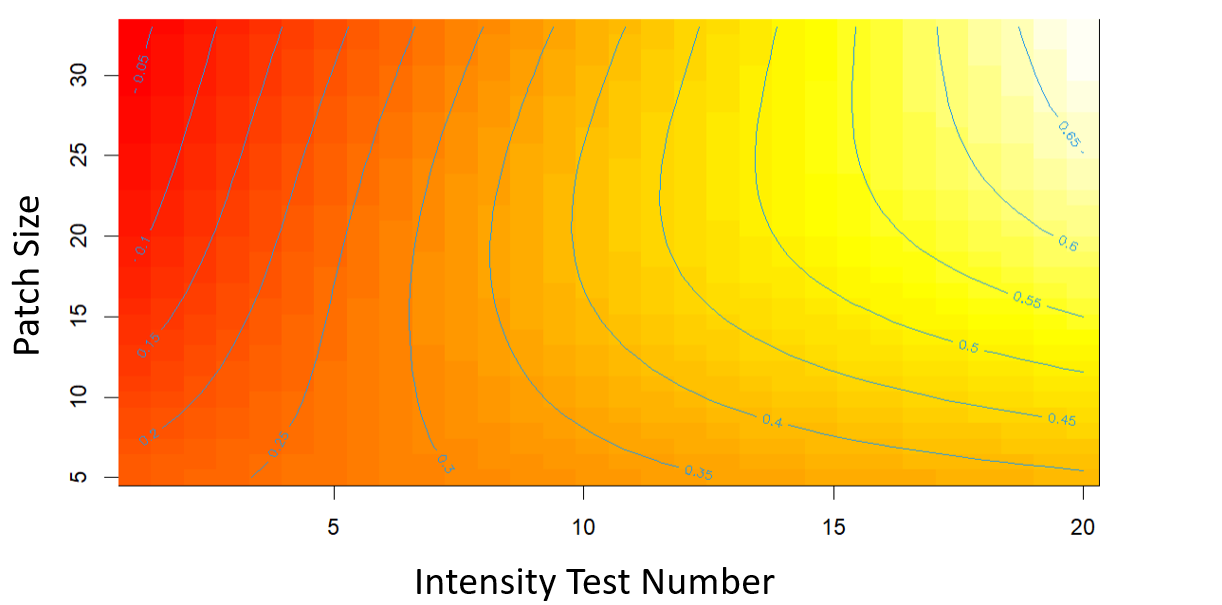}
	\caption{The smoothed accuracy heat map with the score contour.}\label{fig:grad} 
\end{figure}

Figure~\ref{fig:am}a shows the accuracy increases from the bottom left to the top right, and the highest score is $0.80\pm0.09$. However, the result  fluctuates greatly throughout the accuracy map, even though we repeated the experiment.  In particular, the intensity test number $n_d=1,2,3$ showed a random result, especially the unbalanced output along the patch size increment. Therefore, we occupied the generalized adaptive models~\cite{wood2006low} with the tensor product to smooth to find the true appearance of the accuracy map, which is shown in Figure~\ref{fig:grad}b. Both figures' highest scores are concentrated in the top right corner. It indicates that a better classification result can be achieved by the increment of patch size and intensity test number. However, we were limited by the experiment environment and we could not discover the texture classification result when the number of intensity tests was greater than 20.

As the texture classification accuracy is not directly related to the intensity test number or patch size, Kendall's $\tau$~\cite{kendall1938new} is used to offer the correlation analysis between each parameter and the smoothed accuracy, which is listed in Table~\ref{tab:tautest}. It is clear that both parameters are positively and significantly correlated with classification accuracy. However, we could draw the conclusion that the intensity test number has a greater effect on the accuracy than the patch size. The patch size does not influence the classification accuracy much as the correlation coefficient is $0.1109$. Based on this result, we would recommend selecting the intensity test number according to the computational resources with a fixed patch size for the further practical use of BRIEF.
\begin{table}[ht!]
	\centering
	\small
	\caption{Kendall's $\tau$ for Patch Size and Intensity Test Number}
	\label{tab:tautest}
	\begin{tabular}{c|cr}
		\hline\toprule
		%\multirow{2}{*}{} 	&\multicolumn{2}{c}{Parameter} \\
		&$\tau$&p-value\\\midrule
		Patch Size&$0.1109$&$0.005$\\
		Intensity Test Number&$0.8264$&$<0.001$\\
		\bottomrule\hline
	\end{tabular}
\end{table}

\section{Evaluation}\label{sec:evaluation}
To evaluate the performance of C-BRIEF with the optimal selection of parameters, we used the framework of RoTeB~\cite{liu2016evaluation} to compare C-BRIEF to the other methods under three aspects: overall classification accuracy, noise/blur robustness and computational efficiency. Specifically, we tested our methods on Outex\_TC10, 11, and 12~\cite{ojala2002outex}, CUReT~\cite{varma2002classifying}, KTH-TIPS2-b~\cite{caputo2005class}, ALOT~\cite{burghouts2009material}, DTD~\cite{cimpoi14describing}, Brodatz~\cite{brodatz1966textures}, BrodatzRot~\cite{brodatz}, UMD~\cite{xu2009viewpoint} and UIUCTex~\cite{lazebnik2005sparse}.

The optimal parameter selection of C-BRIEF is based on Section~\ref{sec:optimize}. Precisely, for the patch size and intensity test number, we chose 15 and 10 respectively in consideration of the computational resource. Hence, as the result offered by RoTeB contains several descriptor sizes, we selected the result of LBP~\cite{ojala2002multiresolution} and CLBP\_HF~\cite{zhao2011rotation} with the neighbourhood size $11\times 11$ in comparing the dataset Outex\_TC10 \& 12. However, we reproduced the result of BRINT~\cite{liu2014brint} for the whole comparison with the same neighbourhood size as LBP and CLBP\_HF. %Specifically, we did not include MRELBP~\cite{liu2016median}. It is because the minimum sampling patch size of MRELBP is $19\times 19$, which is much larger than C-BRIEF and other methods. 
Classification of the traditional methods was by using the 1-NN classifier for all methods except the DTD, where the support vector machine (SVM)~\cite{platt1999probabilistic} was used. The deep methods selected in RoTeB are the Fisher vector with VGG and AlexNet, which are labelled as FV-CNN and FV-AlexNet, respectively~\cite{cimpoi2015deep}. It is noticed that the classifier used on FV-CNN and FV-AlexNet is SVM.

\begin{table*}[t!]
	\centering
	\small
	\caption{Classification Accuracy (\%) of Different Dataset}
	\label{tab:classresult}
	%\begin{adjustwidth}{-0.7cm}{}
		\begin{tabular}{l|p{0.7cm}p{0.7cm}p{0.7cm}p{0.7cm}p{0.7cm}|p{0.7cm}p{0.7cm}p{0.7cm}p{0.7cm}p{0.7cm}}
			\hline\toprule
			\multirow{2}{*}{\begin{tabular}[c]{@{}l@{}}\\\\\\Method\end{tabular} } &\multicolumn{5}{c|}{Colour} &\multicolumn{5}{c}{Greyscale}                                                                      \\
			
			& \rotatebox[origin=c]{90}{Outex\_TC10}  & \rotatebox[origin=c]{90}{CUReT}& \rotatebox[origin=c]{90}{KTH-TIPS2-b} & \rotatebox[origin=c]{90}{ALOT} & \rotatebox[origin=c]{90}{DTD}& \rotatebox[origin=c]{90}{Outex\_TC12} &   \rotatebox[origin=c]{90}{Brodatz} & \rotatebox[origin=c]{90}{BrodatzRot} & \rotatebox[origin=c]{90}{UMD} & \rotatebox[origin=c]{90}{UIUCTex}  \\
			\midrule
			LBP$^{riu2}$~\cite{ojala2002multiresolution}     & $99.01$ &$97.03$&$62.69$&$94.15$&$37.09$&$89.12$ &$90.70$&$79.22$&$96.15$&$88.36$\\
			CLBP-HF~\cite{zhao2011rotation}  & $\mathbf{99.56}$ &$97.05$&$68.10$&$96.30$&$50.21$&$94.07$ &$91.95$&$82.07$&$97.24$&$92.55$\\
			BRINT\_CS\_CM~\cite{liu2014brint}&$98.50$&$91.67$&$94.14$&$82.03$&$48.33$&$94.00$&$88.00$&$81.00$&$84.00$&$83.33$\\
			FV-CNN (SVM)~\cite{cimpoi2015deep}&$80.00$&$99.00$&$81.80$&$\mathbf{99.50}$&$\mathbf{72.30}$&$82.30$&$\mathbf{98.70}$&$\mathbf{92.10}$&$\mathbf{99.90}$&$\mathbf{99.80}$\\
			FV-AlexNet (SVM)~\cite{cimpoi2015deep}&$67.30$&$98.40$&$77.90$&$99.10$&$62.90$&$72.30$&$98.20$&$83.10$&$99.70$&$99.10$\\\midrule

			C-BRIEF$_{Uniform}$&$97.00$&$\mathbf{99.56}$&$\mathbf{98.60}$&$94.30$&$38.75$&$\mathbf{96.20}$&$88.83$&$79.33$&$93.25$&$82.75$\\
			C-BRIEF$_{Gaussian}$&$82.08$&$88.23$&$95.91$&$83.49$&$34.11$&$50.63$&$66.67$&$61.54$&$88.73$&$80.33$\\
			\bottomrule\hline 
		\end{tabular}
		
	%\end{adjustwidth}
\end{table*}
\begin{table*}[t!]
	\centering
	\small
	%\ContinuedFloat*
	\caption{Classification Accuracy (\%) of Noise and Blur Robustness Based on Outex\_TC11 (24 classes)}
	
	\label{tab:noiseresult11}
	%\begin{adjustwidth}{-0.4cm}{}
		\begin{tabular}{l|p{0.7cm}p{0.7cm}p{0.7cm}p{0.7cm}|p{1.5cm}|p{0.7cm}p{0.7cm}p{0.7cm}p{0.7cm}p{0.7cm}}
			\hline\toprule
			\multirow{2}{*}{\begin{tabular}[c]{@{}l@{}}\\\\\\Method\end{tabular} } &\multicolumn{4}{c|}{Gaussian Blur}&Gaussian Noise   &\multicolumn{5}{c}{Salt and Pepper Noise}                                                                 \\
			
			& \centering\rotatebox[origin=c]{90}{$\sigma=0.5$} & \centering\rotatebox[origin=c]{90}{$\sigma=0.75$} & \centering\rotatebox[origin=c]{90}{$\sigma=1$}& \centering\rotatebox[origin=c]{90}{$\sigma=1.25$} & \centering\rotatebox[origin=c]{90}{$\sigma=5$} & \centering\rotatebox[origin=c]{90}{$\rho=0.05$}&\centering\rotatebox[origin=c]{90}{$\rho=0.15$} & \centering\rotatebox[origin=c]{90}{$\rho=0.3$} & \centering\rotatebox[origin=c]{90}{$\rho=0.4$} & \rotatebox[origin=c]{90}{$\rho=0.5$}  \\
			\midrule
			LBP$^{riu2}$~\cite{ojala2002multiresolution}    & \centering$94.2$ &$46.5$ &$24.6$&$12.7$&\centering$17.7$&\centering$31.7$&\centering$4.2$&\centering$4.2$&\centering$4.4$&$4.2$\\
			CLBP-HF~\cite{zhao2011rotation}  & $99.6$ &$81.3$ &$47.9$&$29.4$&\centering$20.6$&\centering$14.4$&\centering$4.2$&\centering$4.2$&\centering$4.2$&$4.2$\\
			BRINT\_CS\_CM~\cite{liu2014brint}&$90.2$&$73.3$&$38.5$&$24.5$&\centering$91.4$&\centering$\mathbf{64.5}$&\centering$\mathbf{12.9}$&\centering$\mathbf{8.5}$&\centering$5.1$&$3.9$\\
			FV-CNN (SVM)~\cite{cimpoi2015deep}&\centering$\mathbf{100.0}$&$\mathbf{100.0}$&$\mathbf{96.5}$&$\mathbf{89.8}$&\centering$93.1$&\centering$21.0$&\centering$12.1$&\centering$6.0$&\centering$6.5$&$4.2$\\
			FV-AlexNet (SVM)~\cite{cimpoi2015deep}&\centering$\mathbf{100.0}$&$98.8$&$87.7$&$60.4$&\centering$81.5$&\centering$10.4$&\centering$6.7$&\centering$6.6$&\centering$4.2$&$4.2$\\\midrule
			C-BRIEF$_{Uniform}$&$\mathbf{100.0}$&\centering$88.9$&\centering$66.6$&\centering$38.9$&\centering$\mathbf{100.0}$&\centering$5.1$&\centering$5.3$&\centering$4.1$&\centering$4.2$&$4.2$\\
			C-BRIEF$_{Gaussian}$&\centering$99.4$&\centering$73.1$&\centering$44.5$&\centering$31.3$&\centering$99.9$&\centering$6.9$&\centering$4.2$&\centering$4.2$&\centering$\mathbf{7.1}$&$4.2$\\
			\bottomrule\hline 
		\end{tabular}
	%\end{adjustwidth}
\end{table*}
\begin{table*}[ht!]
	\centering
	\small
	%\ContinuedFloat
	\caption{Classification Accuracy (\%) of Noise and Blur Robustness Based on Outex\_TC23 (68 classes)}
	\label{tab:noiseresult23}
	%\begin{adjustwidth}{-0.4cm}{}
		\begin{tabular}{l|p{0.7cm}p{0.7cm}p{0.7cm}p{0.7cm}|p{1.5cm}|p{0.7cm}p{0.7cm}p{0.7cm}p{0.7cm}p{0.7cm}}
			\hline\toprule
			\multirow{2}{*}{\begin{tabular}[c]{@{}l@{}}\\\\\\Method\end{tabular} } &\multicolumn{4}{c|}{Gaussian Blur}&Gaussian Noise   &\multicolumn{5}{c}{Salt and Pepper Noise}                                                                 \\
			
			& \centering\rotatebox[origin=c]{90}{$\sigma=0.5$} & \centering\rotatebox[origin=c]{90}{$\sigma=0.75$} & \centering\rotatebox[origin=c]{90}{$\sigma=1$}& \centering\rotatebox[origin=c]{90}{$\sigma=1.25$} & \centering\rotatebox[origin=c]{90}{$\sigma=5$} & \centering\rotatebox[origin=c]{90}{$\rho=0.05$}&\centering\rotatebox[origin=c]{90}{$\rho=0.15$} & \centering\rotatebox[origin=c]{90}{$\rho=0.3$} & \centering\rotatebox[origin=c]{90}{$\rho=0.4$} & \rotatebox[origin=c]{90}{$\rho=0.5$}  \\
			\midrule
			LBP$^{riu2}$~\cite{ojala2002multiresolution}    & \centering$72.4$ &$30.3$ &$16.6$&\centering$9.7$&\centering$8.4$&\centering$11.8$&\centering$1.5$&\centering$1.5$&\centering$1.5$&$1.5$\\
			CLBP-HF~\cite{zhao2011rotation}  & $85.4$ &$59.2$ &$39.1$&$25.1$&\centering$17.5$&\centering$6.0$&\centering$2.9$&\centering$1.6$&\centering$1.4$&$1.5$\\
			BRINT\_CS\_CM~\cite{liu2014brint}&\centering$92.0$&\centering$65.7$&\centering$39.3$&\centering$29.6$&\centering$48.9$&\centering$\mathbf{23.7}$&\centering$3.0$&\centering$2.1$&\centering$2.8$&$1.5$\\
			FV-CNN (SVM)~\cite{cimpoi2015deep}&\centering$\mathbf{99.6}$&$94.1$&$\mathbf{83.1}$&$\mathbf{71.8}$&\centering$\mathbf{71.5}$&\centering$10.3$&\centering$\mathbf{5.2}$&\centering$2.3$&\centering$1.5$&$1.8$\\
			FV-AlexNet (SVM)~\cite{cimpoi2015deep}&\centering$97.1$&$82.8$&$63.6$&$43.4$&\centering$46.0$&\centering$2.8$&\centering$5.0$&\centering$\mathbf{4.3}$&\centering$\mathbf{3.1}$&$1.5$\\\midrule
			C-BRIEF$_{Uniform}$&$99.3$&$\mathbf{94.6}$&$73.6$&$50.6$&\centering$47.3$&\centering$2.5$&\centering$1.5$&\centering$1.5$&\centering$2.3$&$1.5$\\
			C-BRIEF$_{Gaussian}$&$93.7$&$70.8$&$46.7$&$32.6$&\centering$44.9$&\centering$4.2$&\centering$2.5$&\centering$2.5$&\centering$2.5$&$\mathbf{2.5}$\\
			\bottomrule\hline 
		\end{tabular}
	%\end{adjustwidth}
\end{table*}

\begin{table}[t!]
    \centering
	\small
	%\ContinuedFloat
	\caption{Average Feature Extraction Time for Each Method}
	\label{tab:processtime}
    \begin{tabular}{c|c}
    \hline\toprule
         Descriptor& Feature Extraction Time (ms)\\\midrule
         LBP$^{riu2}$~\cite{ojala2002multiresolution}&87.2\\
         CLBP-HF~\cite{zhao2011rotation}& 256.2\\
         BRINT\_CS\_CM~\cite{liu2014brint}& 771.0\\
         FV-CNN~\cite{cimpoi2015deep}& 2655.4\\
         FV-AlexNet~\cite{cimpoi2015deep}& 238.6\\\midrule
			C-BRIEF$_{Uniform}$&152.2\\
			C-BRIEF$_{Gaussian}$& 152.2\\
			\bottomrule\hline 
    \end{tabular}
\end{table}

%% one of these C-BRIEF$_{Uniform}$ must be C-BRIEF$_{Gaussian} - i've guessed which
From Table~\ref{tab:classresult},the general classification performance of C-BRIEF$_{Uniform}$ is better than C-BRIEF$_{Gaussian}$. C-BRIEF$_{Uniform}$ achieved the highest accuracy in the greyscale dataset, Outex\_TC12, and colour datasets, CUReT and KTH-TIPS2-b, with the classification score of $96.20\%\pm0.9$, $99.56\%\pm0.6$, and $98.60\%\pm0.4$, respectively. These three datasets contain material texture and a limited number of classes. Also, we noticed that the C-BRIEF$_{Gaussian}$ performed poorly compared to C-BRIEF$_{Uniform}$. It might be related to the Gaussian random generator, which needs further investigation. We still can conclude that C-BRIEF$_{Uniform}$ is suitable for representing static texture objects, such as material or medical images.  

Even though the performance of C-BRIEF$_{Uniform}$ is competitive with FV-CNN, the classification score of C-BRIEF$_{Uniform}$ drops largely with DTD. It was because of the various appearances of texture that C-BRIEF$_{Uniform}$ could not offer a stable texture representation. Similarly, LBP$^{riu2}$, BRINT\_CS\_CM and LBP-HF show a drop in the classification score. On the other hand, benefitting from the multiple filters within the convolutional layers, FV-CNN and FV-AlexNet can handle the abstract texture attribute in DTD. They offer robustness and high-level abstraction in texture extraction.

However, dealing with the greyscale datasets, C-BRIEF$_{Uniform}$ achieved similar results as LBP$^{riu2}$, resulting from the degeneration of BRIEF. Also, we found the accuracy drop when C-BRIEF dealt with the large rotation and limited training sample, such as BrodatzRot and Brodatz, respectively.

For the noise robustness test, RoTeB uses 3 types of noise or blur: Gaussian blur (with standard deviation $\sigma=0.5,\ 0.75,\ 1,\ 1.25$), Gaussian noise (with standard deviation $\sigma=5$) and Salt and Pepper noise (with density $\rho=0.5,\ 0.15,\ 0.3,\ 0.4,\ 0.5$). The training sets are noise-free images. Subsequently, the classification score would be given by comparing noisy images' histograms with the normal ones. The classification results based on Outex\_11 and 23 are listed in Table~\ref{tab:noiseresult11}a and b, respectively. C-BRIEF$_{Gaussian}$ still could not achieve better results than C-BRIEF$_{Uniform}$. However, C-BRIEF$_{Uniform}$ offers similar results as FV-CNN under the Gaussian blur and noise. Especially, it achieved $100\%$ classification accuracy in Gaussian blur with $\sigma=0.5$ and the same type of noise. This performance may be a benefit from the noise threshold $\theta^{\prime}$. However, it may also be the reason that C-BRIEF cannot handle the salt and pepper noise well. 

Table~\ref{tab:processtime} shows the average feature extraction time for each method. Specifically, the extraction time for FV-CNN and FV-AlexNet only covers the network's processing time without considering the training time. Compared with other methods, C-BRIEF is the second fastest method in extracting texture features. According to the evaluation of RoTeB, it shows that C-BRIEF is an efficient and powerful texture descriptor in extracting static colour texture, with a certain amount of robustness against  noise and blur.

\section{Conclusion}
In this paper, we analysed the behaviour of each BRIEF parameter under the texture classification problem and offered the colour extended BRIEF (C-BRIEF) for the colour texture classification. We found that the original geometric pattern could be directly used for the texture classification. The stricter noise threshold $\theta^{\prime}$ is recommended to achieve a better classification score. It was found that the patch size did not greatly influence the classification score but the intensity test number did. Moreover,  overloading the intensity tests in a fixed sampling patch  achieved a better classification result. However, it is not clearly seen whether there is an upper bound for the intensity test number, even though the maximum number of the intensity tests of a given $S\times S $ patch is $\nicefrac{S^2(S^2-1)}{2}$. However, Kendall's $\tau$ result showed that the intensity test number could be considered prior to the patch size in the further usage of BRIEF. It is a different behaviour between BRIEF and LBP and the related methods, that once you have decided the neighbourhood size, the pattern and intensity test numbers are fixed.

Applying the optimal parameters on the C-BRIEF, we evaluated the method by using the framework of RoTeB. The result showed that C-BRIEF's performance on the colour texture dataset was competitive with the FV-CNN. However, as the degenerate of C-BRIEF on the greyscale, it only achieved the best score on Outex\_TC12. For the noise and blur robustness, C-BRIEF performed  excellently against Gaussian noise and blur. However, facing salt and pepper noise, it did not outperform the other hand-crafted descriptors. Compared to LBP and the related methods, BRIEF and C-BRIEF can be flexible with the patch size without considering the symmetric pattern, resulting in the odd neighbourhood size. In general, the proposed colour descriptor, C-BRIEF, is efficient and powerful in extracting the static texture feature.  

\bibliographystyle{ieeetr}
\bibliography{egbib.bib}

\appendix
\begin{table}[t!]
	\centering
	\small
	\caption{Group Number of Texture Combination. The category number of Direction column: 1. Directed \textit{vs.} Directed; 2. Directed \textit{vs.} Undirected (including Undirected \textit{vs.} Directed); 3. Undirected \textit{vs.} Undirected.}
	\label{tab:combination}
	\begin{tabular}{lllc}
		\hline\toprule
		No.&Description& &Direction\\ 
		\midrule
		1&T01\_bark1&\& \  T02\_bark2&1\\
		2&T01\_bark1&\& \  T08\_granite&2\\
		3&T01\_bark1&\& \ T10\_floor1&2\\
		4&T01\_bark1&\& \  T16\_glass1&1\\
		5&T01\_bark1&\& \  T19\_carpet2&2\\
		6&T01\_bark1&\& \  T24\_corduroy&1\\
		7&T02\_bark2&\& \  T08\_granite&2\\
		8&T02\_bark2&\& \  T10\_floor1&2\\
		9&T02\_bark2&\& \  T16\_glass1&1\\
		10&T02\_bark2&\& \  T19\_carpet2&2\\
		11&T02\_bark2&\& \ T24\_corduroy&1\\
		12&T08\_granite&\& \  T10\_floor1&3\\
		13&T08\_granite&\& \  T16\_glass1&2\\
		14&T08\_granite&\& \  T19\_carpet2&3\\
		15&T08\_granite&\& \  T24\_corduroy&2\\
		16&T10\_floor1&\& \  T16\_glass1&2\\
		17&T10\_floor1&\& \ T19\_carpet2&3\\
		18&T10\_floor1&\& \ T24\_corduroy&2\\
		19&T16\_glass1&\& \ T19\_carpet2&2\\
		20&T16\_glass1&\& \ T24\_corduroy&1\\
		21&T19\_carpet2&\& \ T24\_corduroy&2\\
		\bottomrule\hline
	\end{tabular}
\end{table}

\iffalse
\begin{figure*}[t!]
	\centering
	\small
	\includegraphics[width = 2\columnwidth]{Chi_squared_sum.jpg}
	\caption{Different distribution accuracy rate ($n_d=9, S=5, N=1$) of pixel-wise classification.}
	\label{fig:distribution}
\end{figure*}\fi
\end{document}